\pdfoutput=1

\documentclass[11pt]{article}

\usepackage[]{acl}

\usepackage{times}
\usepackage{latexsym}

\usepackage[T1]{fontenc}

\usepackage[utf8]{inputenc}

\usepackage{microtype}
\usepackage{bm}
\usepackage{subfigure}
\usepackage{graphicx}
\usepackage{multirow}
\usepackage{amsmath}
\usepackage{adjustbox}
\usepackage{array}
\usepackage{CJKutf8}
\usepackage{makecell}

\usepackage{booktabs}

\definecolor{darkgreen}{rgb}{0,1.0,0}

\title{Improving Lexical Embeddings for Robust Question Answering}
\author{Weiwen Xu\raisebox{4pt}{\small $1,2$}~\thanks{~~Work was done when the author was a staff in Institute for Infocomm Research, A*STAR.}, Bowei Zou\raisebox{4pt}{\small $1$}, Wai Lam\raisebox{4pt}{\small $2$}, Ai Ti Aw\raisebox{4pt}{\small $1$}\\
  \raisebox{4pt}{\small $1$}Institute for Infocomm Research, A*STAR\\
  \raisebox{4pt}{\small $2$}The Chinese University of Hong Kong\\
    {\tt\{wwxu,wlam\}@se.cuhk.edu.hk}\qquad  \\{\tt \{zou\_bowei,aaiti\}@i2r.a-star.edu.sg}}
\begin{document}
\maketitle
\begin{abstract}
Recent techniques in Question Answering (QA) have gained remarkable performance improvement with some QA models even surpassed human performance. However, the ability of these models in truly understanding the language still remains dubious and the models are revealing limitations when facing adversarial examples. To strengthen the robustness of QA models and their generalization ability, we propose a representation Enhancement via Semantic and Context constraints (ESC) approach to improve the robustness of lexical embeddings. Specifically, we insert perturbations with semantic constraints and train enhanced contextual representations via a context-constraint loss to better distinguish the context clues for the correct answer. Experimental results show that our approach gains significant robustness improvement on four adversarial test sets.


\end{abstract}

\section{Introduction}
With the help of recent developments in neural networks~\cite{seo2016bidirectional, devlin-etal-2019-bert} as well as the release of high-quality datasets~\cite{rajpurkar-etal-2016-squad, joshi2017triviaqa, rajpurkar2018know}, the QA task has made unprecedented progress. On the famous SQuAD dataset, QA models have achieved higher performance than human. However, the high performances of existing models do not account for the full understanding of language, as models often rely on recognizing the predictive patterns on test sets without learning the deep semantics beneath the language~\cite{rimell-etal-2009-unbounded, paperno-etal-2016-lambada}.

\begin{figure}
    \centering
     \includegraphics[scale=0.6]{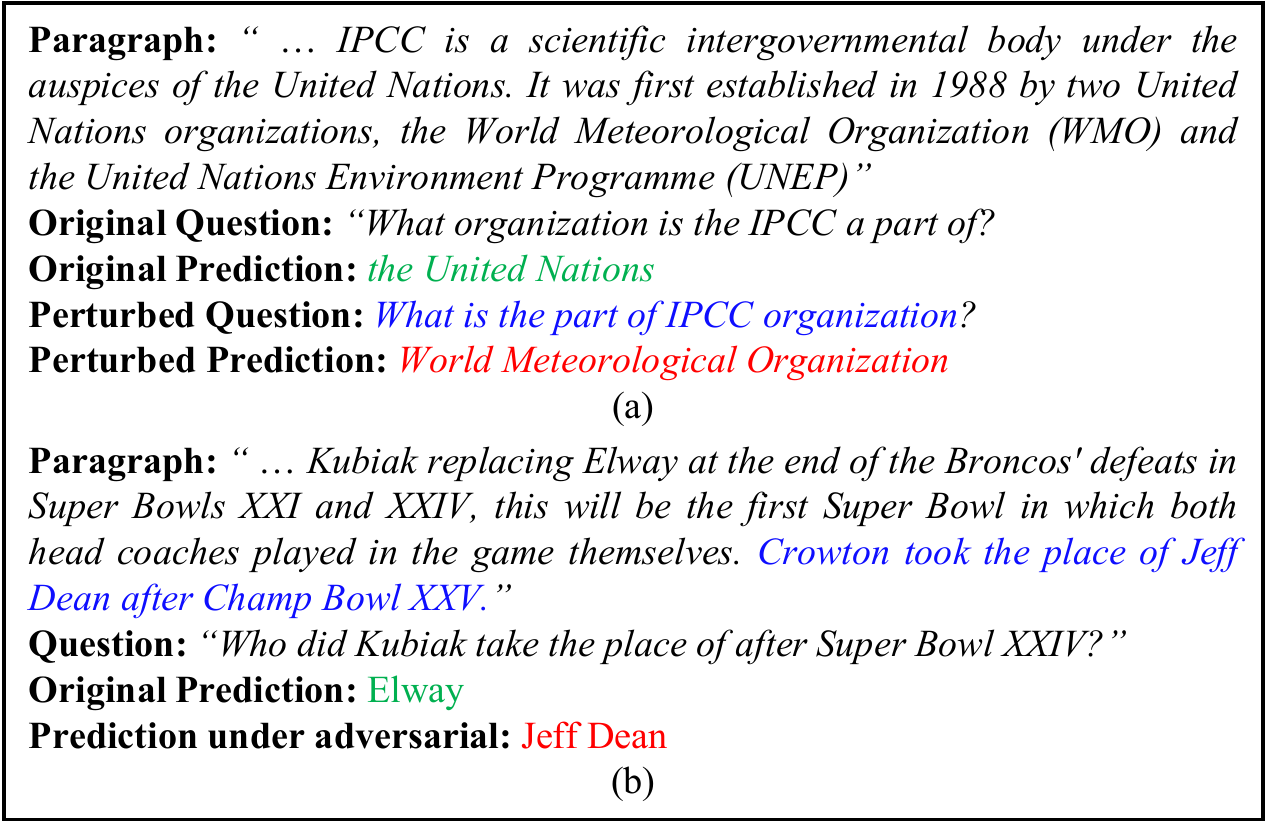} 
       \caption{Examples of oversensitivity (a) and overstability (b). BERT model gets fooled by the additional adversarial input ({\color{blue} in blue}).}
    \label{fig:case_example}
    \vspace{-16pt}
\end{figure}

As shown in Figure~\ref{fig:case_example}(a), a small perturbation in the question may mislead the model to get the wrong answer. This issue is presented as oversensitivity~\cite{szegedy2013intriguing, goodfellow2014explaining}, causing models incapable to generalize to semantically invariant text perturbations. 
Figure~\ref{fig:case_example}(b) shows another scenario, where adding one distracting sentence 
sharing sufficient words with the question but meaningfully changed, is enough to confuse the model to select the distractor as the answer. This issue is described as overstability~\cite{jia-liang-2017-adversarial}. 
Both issues are difficult to be addressed by using superficial lexical representations (e.g., simple word embeddings) which is not always enough for QA models to distinguish the texts containing the correct answer and the distracting context. 
Existing methods solve these robustness issues mainly by refining kinds of attentions to make models predict precisely~\cite{huang2017fusionnet, hu2017reinforced,xu-etal-2021-addressing}, adversarial training based methods~\cite{miyato2016adversarial, alzantot2018generating, iyyer-etal-2018-adversarial}, learning the deep mutual information among the paragraph, the question and its answer~\cite{yeh-chen-2019-qainfomax, xu-etal-2021-dynamic}, or using external knowledge to build adversarial examples to enlarge the training size~\cite{wang-bansal-2018-robust, zhou2020robust}.

This paper addresses the robustness issues by improving the robustness of lexical embeddings to make them less sensitive to variations. 
To this end, we propose an ESC approach: 
1) We insert perturbations with a semantic constraint to the input paragraphs or questions, which makes QA models see more semantically related samples. 
2) We enhance contextual representations by minimizing the context-constraint loss, to stabilize the contextual representations in the presence of perturbation.
We expect the context constraint can precisely represent the context clues to facilitate QA models to distinguish the correct sentence against the distracting one. Experimental results show that the proposed ESC approach gains significant robustness improvement on four adversarial test sets -- \textsc{AddSent}, \textsc{AddOneSent}~\cite{jia-liang-2017-adversarial}, and \emph{Para-Q}, \emph{Adv-Q}~\cite{gan-ng-2019-improving}, yielding absolute F1 improvements of 4.1, 3.7, and 1.0, 4.2 over the BERT baseline, respectively. 

\begin{figure}
    \centering
     \includegraphics[scale=0.6]{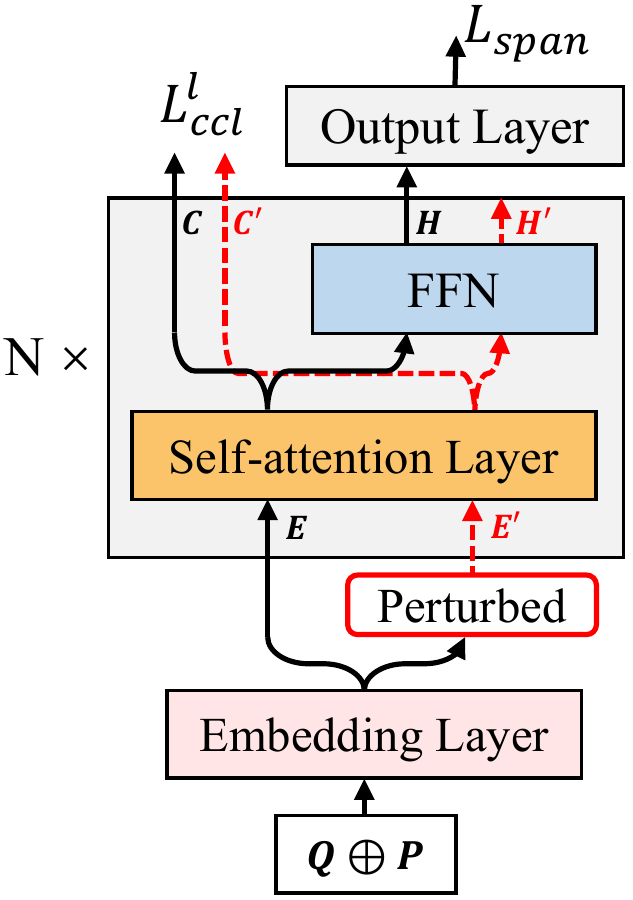} 
       \caption{ The architecture of ESC. The black solid lines indicate the original path of BERT, while the red dotted lines for adversarial input.}
    \label{fig:model}
    \vspace{-15pt}
\end{figure}

\section{Approach}\label{sec:method}

The extractive QA task aims to find an answer span ${\bm A} = \{a_1,..,a_M\}$ inside the paragraph ${\bm P} = \{p_1,..,p_N\}$, which can correctly answer the question ${\bm Q} = \{q_1,..,q_K\}$. We implement ESC over the pretrained BERT~\cite{devlin-etal-2019-bert} for its great superiority on recent Natural Language Processing (NLP) tasks.

As shown by the black solid line in Figure~\ref{fig:model}, BERT gets input of the concatenation of question and paragraph, and maps it into a sequence of word embeddings: ${\bm E}=\{e^q_1,..,e^q_K,e^p_1,..,e^p_N \}$, where $e^q$ and $e^p$ are the word embeddings of question and paragraph, respectively. Then a self-attention mechanism~\cite{Transformer} is introduced to attend all the words of the input. Such self-attention is the only mechanism that allows the interaction between the current word and all surrounding words regardless of their distances. Therefore, we refer the output of self-attention as \emph{contextual representation}: ${\bm C}=\{c^q_1,..,c^q_K,c^p_1,..,c^p_N \}$, which is then projected together with the word embeddings to produce the hidden state ${\bm H}$. Based on the hidden state, we find the most likely span to answer the question.

\subsection{Perturbations with Semantic Constraint}
\label{sec:semantic_const}


To address the sensitivity of non-robust QA models to the trivial perturbations in the input question, we follow the robust training strategies~\cite{belinkov2017synthetic, cheng-etal-2018-towards} but insert perturbations with semantic constraint, as such perturbations would potentially improve the generalization ability of models by slightly changing semantics of contexts during the training step. 

As depicted in Figure~\ref{fig:model}, we automatically insert perturbations with semantic constraint in the input sequence to generate an adversarial sample. Specifically, we randomly select multiple positions in each input sequence with a probability $\sigma$ and perturb the corresponding words. For the word $w_i$ chosen to be perturbed, we build a dynamic set $\mathcal{V}_{w_i}$ consisting of \textit{n} words that have the highest cosine similarities with $w_i$ (exclude $w_i$). We average the embeddings of the words in $\mathcal{V}_{w_i}$ to construct the perturbation for $w_i$. 
\vspace{-5pt}
\begin{equation}
    \mathcal{V}_{w_i} = \mathop{top\_n}\limits_{w_j \in \mathcal{D}, j \ne i} (cos(e_i, e_j))
\end{equation}
\begin{equation}
    e_i' = \frac{1}{n}\sum_{ w_j \in  \mathcal{V}_{w_i}}e_j
\end{equation}
where $\mathcal{D}$ is the dictionary, $e_i$ is the word embedding for $w_i$ and $e_i'$ is the perturbation for $w_i$. We perturb $w_i$ by replacing $e_i$ with $e_i'$.
\subsection{Representations with Context Constraint}

On the other hand, as \citet{yeh-chen-2019-qainfomax} argued, QA models tend to choose a span after the word that existing in the question. We find that the contextual representations fail to extract decisive clues to determine the answer, which causes the model to apply a simple pattern-match policy. Therefore, we utilize the adversarial samples generated with semantic constraint in Sec.~\ref{sec:semantic_const} to teach the model how to react to the context change and thus learn the contextual representations to better represent the context clues. As shown in Figure~\ref{fig:model}, the original input and the adversarial sample go parallelly in ESC. We keep the contextual representations of the adversarial sample as similar as those of the original input by a Context Constraint Loss (CCL), which teaches ESC to pay less attention to the word itself and retain most context clues of the current position.

Formally, let ${\bm C}_l$ and ${\bm C}_l'$ be the output of self-attention in the $l$-th BERT layer for the original input and the adversarial sample respectively. The CCL for each layer is
\vspace{-5pt}
\begin{equation}
\label{loss_l}
\mathcal{L}^l_{ccl} = ||{\bm C}_l - {\bm C}'_l||^2
\end{equation}

Note that the above two constraints are mutually beneficial. On the one hand, the semantic constraint creates an adversarial sample, which works as a teacher to guide the contextual representations learning. On the other hand, the context constraint provides the supervision (${\bm C}_l$) for the adversarial sample, preventing it from shifting too much. We demonstrate it in Table~\ref{tab:ablation}.

\subsection{Training}
Both the original input and the adversarial sample go parallelly in our ESC model. At each layer, we calculate the CCL using ${\bm C}_l$ and ${\bm C}_l'$. We finally use the last layer hidden state of the original input to predict the answer span (See Figure~\ref{fig:model}).

As there are $12/24$ (base/large) layers in BERT, to prevent the CCLs slowing down the training speed and overfitting, we only sample $k$ ($k$ $<$ $12/24$) CCLs to update in each iteration. The final CCL can be written as

\begin{equation}
\label{loss_rel}
\mathcal{L}_{ccl} = \frac{1}{k}\sum_{l \in S} \mathcal{L}^l_{ccl},
\end{equation}
where $S$ contains $k$ sampled layer indexes to be updated.

The final training objective $\mathcal{L}$ is the combination of QA answer span prediction loss $\mathcal{L}_{span}$ and CCL. $\lambda$ is the weighting parameter for CCL.
\vspace{-5pt}
\begin{equation}
\label{loss}
\mathcal{L} = \mathcal{L}_{span} +\lambda\mathcal{L}_{ccl}
\end{equation}

\section{Experimental Settings}
We evaluate the effectiveness of ESC on four challenging adversarial test sets -- \textsc{AddSent}, \textsc{AddOneSent}, \emph{Para-Q} and \emph{Adv-Q}.
The two adversarial test sets, \textsc{AddSent} and \textsc{AddOneSent}, are constructed by~\citet{jia-liang-2017-adversarial} based on SQuAD. In \textsc{AddSent}, they add an adversarial sentence to each paragraph that makes QA models give the worst answer. In \textsc{AddOneSent}, only a randomly adversarial sentence is added. These two are to evaluate the model's robustness on adversarial attack.

\citet{gan-ng-2019-improving} construct another two adversarial test sets: \emph{Para-Q} and \emph{Adv-Q}. In \emph{Para-Q}, they use a paraphrasing model to paraphrase the questions based on PPDB~\cite{pavlick-etal-2015-ppdb}. While in \emph{Adv-Q}, they manually paraphrase the questions. These two are to evaluate the model's ability on generalization.


As \textsc{AddSent} and \textsc{AddOneSent} add distracting sentences in paragraphs and \emph{Para-Q} and \emph{Adv-Q} paraphrase only the questions, we build three variants of our model, where the ESC approach is added to paragraph words (ESC-P), question words (ESC-Q), and  both words (ESC). Detailed implementations can be found in the Appendix~\ref{sec:app1}.

\section{Results and Analysis}
\begin{table}[]
    \small
    \centering
    \setlength{\tabcolsep}{1mm}{
    \begin{tabular}{l|ccc}
    \toprule
       Model & Original  & \textsc{AddSent} & \textsc{AddOneSent}  \\ \hline  \multicolumn{4}{c}{\em Exisiting Approaches} \\ \hline
    QANet        & 83.8 & 45.2 & 55.7 \\
    GQA          & 83.7 & 47.3 & 57.8 \\
    FusionNet    & 83.6 & 51.4 & 60.7 \\
    QAInfomax (B)    & 88.6 & 54.5 & 64.9 \\
    \citeauthor{liu2020robust} &90.4 &63.1& -\\
    AT/VAT (L)  & 92.4 & 63.5 & 72.5 \\ \hline
        \multicolumn{4}{c}{\em Our Approaches} \\ \hline
       BERT (B)     & 87.3 & 52.9 & 63.4 \\
        + ESC-P        & 88.4 & 55.2 & 65.5\\ 
        + ESC-Q        & 87.4 & 54.4 & 65.0\\ 
        + ESC          & \textbf{88.5} & \textbf{57.0} & \textbf{67.1}\\ \hline
        BERT (L) & \textbf{93.3} & 67.8 & 76.5\\
          + ESC-P      & 92.9 & \textbf{69.9} & \textbf{78.3}\\
          + ESC-Q      & 92.3 &69.1 & 76.1\\
          + ESC        & 91.8 & 69.8 & 76.9\\\bottomrule
    \end{tabular}}
    \caption{F1 scores on three test sets (B refers to BERT base model; L: refers to BERT large model). Original is the original SQuAD dev set. Highest scores are bolded.}
    \label{tab:addsent}
    \vspace{-10pt}
\end{table}

Table~\ref{tab:addsent} shows the results on \textsc{AddSent}, \textsc{AddOneSent}, and the original development set. We implement our ESC on strong baselines BERT(B) and BERT(L) and compare with the approaches without using extra data, including QANet~\cite{yu2018qanet}, GQA~\cite{lewis2018generative}, FusionNet~\cite{huang2017fusionnet}, QAInfomax~\cite{yeh-chen-2019-qainfomax}, AT+VAT~\cite{yang2019improving}, and \citet{liu2020robust} to make the improvements of our ESC approach more convincing.\footnote{For a fair comparison, 
we do not compare with \citet{wang-bansal-2018-robust, zhou2020robust, wang2021infobert}, as their works rely on external knowledge in addition to the original SQuAD training set according to our settings.}

\textbf{Performance against Adversarial Attack} As shown in Table~\ref{tab:addsent}, we further improve the performance on both adversarial test sets with the improvements of 4.1 and 3.7 of F1 against the BERT(B), respectively. 
All three variants improve over the baseline BERT significantly, while ESC-P is better than ESC-Q on these test sets. 
The results are within expected as ESC-P is specifically designed to handle and evaluate the performance of the model in this scenario where an adversarial sentence is present in the paragraph texts. By extending the model to handle adversarial attack on the question, ESC still exceeds or maintains its performance on both \textsc{AddSent} and \textsc{AddOneSent}.

\textbf{Performance against Paraphrasing} Column $2^{nd}$ and $3^{rd}$ of Table~\ref{tab:oversen} show the results on \emph{Para-Q} with its original test set \emph{Para-Orig}.\footnote{Detailed EM scores can be found in Appendix\ref{sec:app2}} We compare ESC with the latest approaches: DrQA~\cite{chen2017reading}, BiDAF~\cite{seo2016bidirectional}, and the baseline BERT. The results indicate that 
ESC improves the robustness on adversarial test sets with 1.0 and 0.5 of F1 on BERT(L) and BERT (B), respectively. 
Column $4^{th}$ and $5^{th}$ of Table~\ref{tab:oversen} show the results on \emph{Adv-Q} with its original test set \emph{Adv-Orig}. Since \emph{Adv-Q} only contains 56 samples, the performance on this test set may not fully reflect the model's capability to deal with question paraphrasing. 
We can see that the performance fluctuates since the test set is small. Nevertheless, both results on BERT (B) and BERT (L) show that overall ESC is effective and improves the baseline by 4.2 and 1.3 F1 respectively. The results demonstrate that enhancing the lexical embeddings in both paragraph and question texts helps to improve the robustness of the question answering.

\begin{table}[]
    \centering
    \small
    \begin{tabular}{l|cc|cc}
    \toprule

    Model &\emph{Para-Orig}  & \emph{Para-Q} &\emph{Adv-Orig}  & \emph{Adv-Q}\\ \hline
       \multicolumn{5}{c}{\em Exisiting Approaches} \\ \hline
       DrQA  & 76.3 & 74.3 & 81.0 & 48.9 \\
       BiDAF & 76.9 & 73.5 & 81.6 & 38.3\\ \hline 
            \multicolumn{5}{c}{\em Our Approaches} \\ \hline
       BERT (B)   & 88.0 & 85.3 & 88.9 & 56.6\\ 
       ~~~ +ESC-P    & 87.9 & 85.8 & \textbf{90.3} & 59.6\\ 
       ~~~ +ESC-Q    & 88.0 & 86.2 & 89.4 & \textbf{65.5}\\ 
       ~~~ +ESC      &\textbf{88.9} & \textbf{86.3} & 88.5 & 60.8\\ \hline
       BERT (L)  & 93.3 & 91.0 & 92.5 & 79.6 \\ 
       ~~~ +ESC-P    & 93.1 & 91.3 & 91.2 & 79.3\\ 
       ~~~ +ESC-Q    & \textbf{93.6} &91.3 & 91.1 & 78.0\\ 
       ~~~ +ESC      & 93.2 & \textbf{91.5} & \textbf{92.8} & \textbf{80.9}\\ \bottomrule
    \end{tabular}
    \caption{F1 scores on \emph{Para-Q} and \emph{Adv-Q} and their corresponding original test sets.}
    \label{tab:oversen}
    \vspace{-10pt}
\end{table}

\textbf{Ablation Study} We conduct an ablation study on the two constraints described in Sec.~\ref{sec:method}. Table~\ref{tab:ablation} shows that both constraints improve over the BERT baseline independently, where \textit{+Semantic} performs better on \emph{Para-Q}, and \textit{+Context} performs better on \textsc{AddSent} and \textsc{AddOneSent}. This is in line with our motivation to use semantic and context constraints to address the robustness issue. We also notice that \textit{+Both} gains more improvement than \textit{+Semantic} or \textit{+Context} especially on \textsc{AddSent} and \textsc{AddOneSent}. 
It implies that lexical embeddings and contextual representation affect each other and improving their robustness bring greater benefit and synergy to QA models. 

\begin{table}[]
    \small
    \centering
    \setlength{\tabcolsep}{1mm}{
    \begin{tabular}{l|cccc}
    \toprule
       Model   & \textsc{AddSent} & \textsc{AddOneSent} & Para-Q \\ \hline 
       BERT          & 52.9       & 63.4       & 85.3 \\
       +Semantic     & 53.0 (0.1) & 63.7 (0.3) & 85.9 (0.6)\\ 
       +Context      & 54.4 (1.5) & 64.9 (1.5) & 85.7 (0.4)\\ 
       +Both (ESC)   & 57.0 (4.1) & 67.1 (3.7) & 86.3 (1.0)\\ \bottomrule
    \end{tabular}}
    \caption{Ablation study on three test sets when adding the semantic constraint (+Semantic), the context constraint (+Context) and both constraints (+Both). Improvements against BERT are shown in brackets.}
    \label{tab:ablation}

\end{table}

\begin{figure}
    \centering
     \includegraphics[scale=0.45]{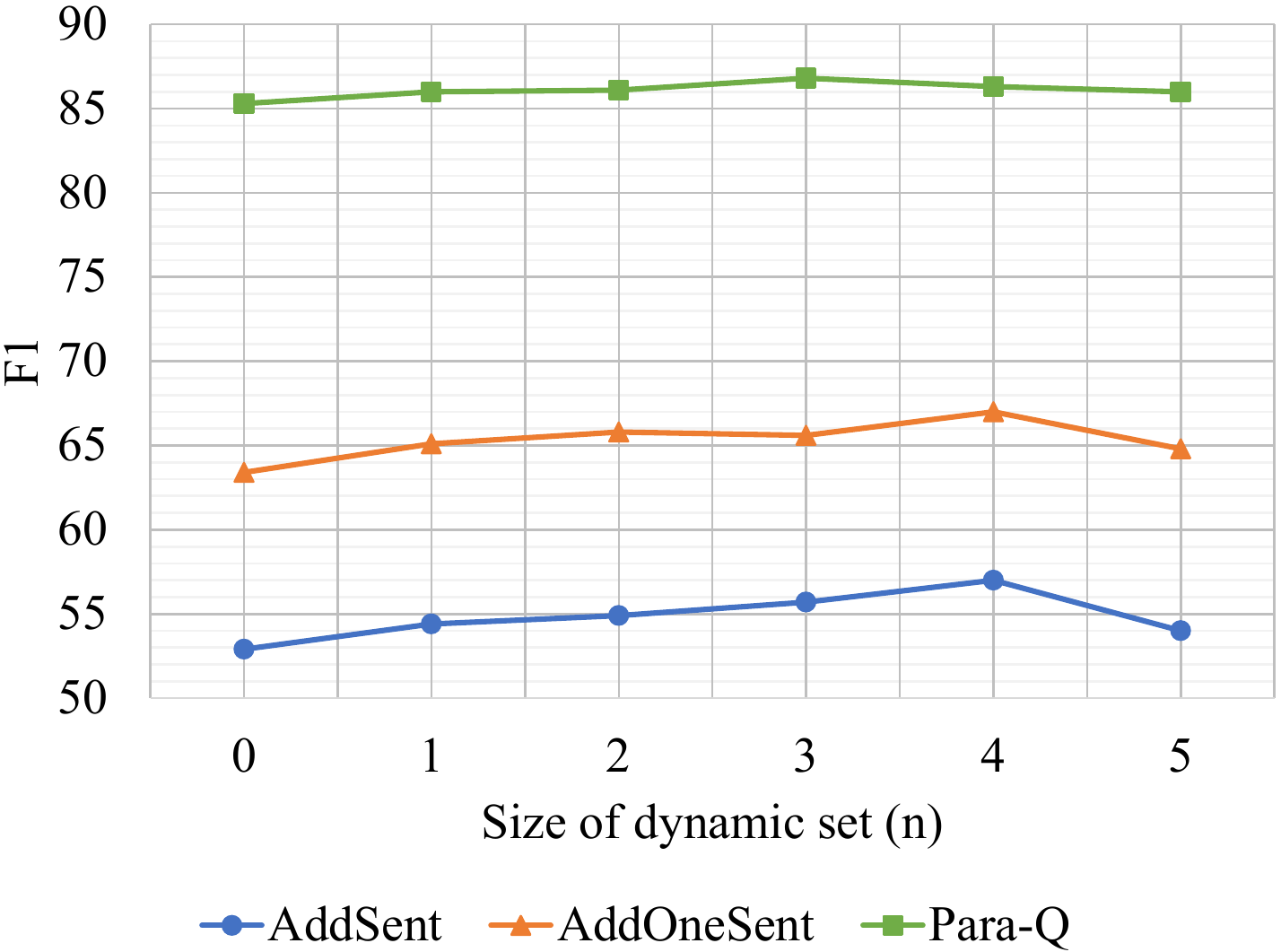} 
       \caption{F1 scores with different dynamic set sizes. Size 0 refers to BERT baseline.}
    \label{fig:dynamic_set}
    \vspace{-10pt}
\end{figure}

\textbf{Effect of Dynamic Set Size} We look into the dynamic set size $n$ to see how it affects the model stability in Figure~\ref{fig:dynamic_set}. The curves are quite stable and consistent where a size of around 4 achieves the best performance among all three test sets.

\section{Conclusion}

This paper presents a novel approach, ESC, to address the robustness issues for QA models. 
We insert perturbations with a semantic constraint to improve the lexical embeddings and enhance context representations via a context constraint loss to make the model more generalizable. 
To verify the effectiveness of our approach, we compare it with strong approaches on four challenging adversarial test sets. 
Experimental results show that ESC can significantly improve the robustness of QA models with regards to adversarial sentences in paragraph and question paraphrasing.

\bibliography{anthology,custom}
\bibliographystyle{acl_natbib}

\appendix
\clearpage
\section{Appendix}
\subsection{Hyper-parameter Settings}
\label{sec:app1}
We use PyTorch re-implementation of BERT\footnote{Available at https://github.com/huggingface/transformers.}. The BERT related hyper-parameter setting remains the same as released for SQuAD1.1 training. For other new hyper-parameters, we perform grid search to find optimal values, where $\sigma$ is selected from $\{0.05, \textbf{0.1}, 0.2\}$, $\lambda$ is selected from $\{0.1, 0.5, \textbf{1}, 2\}$, $n$ is selected from $\{1, 2, 3, \textbf{4}, 5\}$ and $k$ is selected from $\{1, 2, \textbf{3}, 4, 5\}$. The learning rate is set to $3e-5$ and ADAM~\cite{kingma2014adam} optimizer is used for parameters optimization with $\beta_1 = 0.9$ and $\beta_2 = 0.999$. We apply ESC approach both on BERT$_{\text{\tt Base}}$ and BERT$_{\text{\tt Large}}$ backbones. Each ${\tt Base}$ model is trained on one GeForce GTX 1080 Ti card and it takes 10 hours to complete 3 epochs. Each ${\tt Large}$ model is trained on one TITAN RTX card and it takes 8 hours to complete 3 epochs with fp16 enabled.

\subsection{Detailed Results on \emph{Para-Q} and \emph{Adv-Q}}
\label{sec:app2}
In addition to Table~\ref{tab:oversen}, we show the EM scores along with the F1 scores of different approaches on \emph{Para-Q} (Table~\ref{tab:para}) and \emph{Adv-Q} (Table~\ref{tab:adv}) in this section.

\vspace*{12\baselineskip} 
\begin{table}[]
    \centering
    \small
    \begin{tabular}{l|cc|cc}
    \toprule
      \multirow{2}{*}{\textbf{Model}}   & \multicolumn{2}{c|}{EM}  & \multicolumn{2}{c}{F1} \\ \cline{2-5}
       &\emph{Orig-Q}  & \emph{Para-Q} &\emph{Orig-Q}  & \emph{Para-Q}\\ \hline
       \multicolumn{5}{c}{\em Exisiting Approaches} \\ \hline
       DrQA & 67.3 & 65.3 & 76.3 & 74.3 \\
       BiDAF & 67.8 & 63.8 & 76.9 & 73.5 \\ \hline 
            \multicolumn{5}{c}{\em Our Approaches} \\ \hline
       BERT (B) & 80.4 & 76.8 & 88.0 & 85.3\\ 
       ~~~ +ESC-P & 80.0 & 77.9 & 87.9 & 85.8\\ 
       ~~~ +ESC-Q & 80.2 & 78.4 & 88.0 & 86.2\\ 
       ~~~ +ESC & \textbf{82.0} & \textbf{78.5} & \textbf{88.9} & \textbf{86.3}\\ \hline
       BERT (L) & 86.6 & 83.8 & 93.3 & 91.0 \\ 
       ~~~ +ESC-P & 86.8 & 84.6 & 93.1 & 91.3\\ 
       ~~~ +ESC-Q & 88.3 & \textbf{86.0} & \textbf{93.6} &91.3\\ 
       ~~~ +ESC & \textbf{87.7} & 85.3 & 93.2 & \textbf{91.5}\\ \bottomrule
    \end{tabular}
    \caption{EM and F1 scores  on the original questions (\emph{Orig-Q}) compared to the paraphrased questions (\emph{Para-Q}).}
    \label{tab:para}
\end{table}

\begin{table}[]
    \centering
    \small
    \begin{tabular}{c|cc|cc}
    \toprule
      \multirow{2}{*}{\textbf{Model}}   & \multicolumn{2}{c|}{EM}  & \multicolumn{2}{c}{F1} \\ \cline{2-5}
       &\emph{Orig-Q}  & \emph{Adv-Q} &\emph{Orig-Q}  & \emph{Adv-Q}\\ \hline
       \multicolumn{5}{c}{\em Exisiting Approaches} \\ \hline
       DrQA & 71.4 & 39.3 & 81.0 & 48.9 \\
       BiDAF & 75.0 & 30.4 & 81.6 & 38.3 \\ \hline 
            \multicolumn{5}{c}{\em Our Approaches} \\ \hline
       BERT-S (B) & 82.1 & 51.8 & 88.9 & 56.6\\ 
       ~~~ +ESC-P & \textbf{83.9} & 53.5 & \textbf{90.3} & 59.6\\ 
       ~~~ +ESC-Q & 82.1 & \textbf{58.9} & 89.4 & \textbf{65.5}\\ 
       ~~~ +ESC & 82.1 & 53.6 & 88.5 & 60.8\\ \hline
       BERT-S (L) & 87.5 & 75.0 & 92.5 & 79.6 \\ 
       ~~~ +ESC-P & 87.5 & 75.0 & 91.2 & 79.3\\ 
       ~~~ +ESC-Q & 85.7 & 73.2 & 91.1 & 78.0\\ 
       ~~~ +ESC & \textbf{87.6} & \textbf{75.0} & \textbf{92.8} & \textbf{80.9}\\ \bottomrule
    \end{tabular}
    \caption{EM and F1 scores on the original questions (\emph{Orig-Q}) compared to the adversarial paraphrased questions (\emph{Adv-Q}).}
    \label{tab:adv}
    \vspace{-10pt}
\end{table}

\end{document}